\documentclass[runningheads]{llncs}
\usepackage[T1]{fontenc}
\usepackage{graphicx}
\usepackage{amsmath,graphicx,hyperref}
\usepackage{amssymb}
\usepackage{multirow}
\usepackage{color}

\begin{document}
\title{TAFG-MAN: Timestep-Adaptive Frequency-Gated Latent Diffusion for Efficient and High-Quality Low-Dose CT Image Denoising}
\titlerunning{TAFG-MAN}
%
%

\author{Tangtangfang Fang\inst{1}$^*$ \and
Yang Jiao\inst{1}$^*$ \and
Xiangjian He\inst{1}$^\dagger$ \and
Jingxi Hu\inst{1}$^\dagger$ \and
Jiaqi Yang\inst{1}}

\authorrunning{T. Fang, Y. Jiao et al.}
%

\institute{
University of Nottingham Ningbo China, Ningbo, Zhejiang, China\\
\email{tangtangfang\_fang@brown.edu, bixyj13@nottingham.edu.cn,\\Sean.He@nottingham.edu.cn, Jingxi.HU@nottingham.edu.cn,  Jiaqi.YANG2@nottingham.edu.cn}\\
$^*$ These authors contributed equally. $^\dagger$ Corresponding author.
}

\maketitle              
\begin{abstract}
Low-dose computed tomography (LDCT) reduces radiation exposure but also introduces substantial noise and structural degradation, making it difficult to suppress noise without erasing subtle anatomical details. In this paper, we present \textbf{TAFG-MAN}, a latent diffusion framework for efficient and high-quality LDCT image denoising. The framework combines a perceptually optimized autoencoder, conditional latent diffusion restoration in a compact latent space, and a lightweight \textbf{T}imestep-\textbf{A}daptive \textbf{F}requency-\textbf{G}ated (TAFG) conditioning design. TAFG decomposes condition features into low- and high-frequency components, predicts timestep-adaptive gates from the current denoising feature and timestep embedding, and progressively releases high-frequency guidance in later denoising stages before cross-attention. In this way, the model relies more on stable structural guidance at early reverse steps and introduces fine details more cautiously as denoising proceeds, improving the balance between noise suppression and detail preservation. Experiments show that TAFG-MAN achieves a favorable quality-efficiency trade-off against representative baselines. Compared with its base variant without TAFG, it further improves detail preservation and perceptual quality while maintaining essentially the same inference cost, and ablation results confirm the effectiveness of the proposed conditioning mechanism.

\end{abstract}

\section{Introduction}
\label{sec:intro}
Computed tomography (CT) is widely used in clinical practice, but its diagnostic benefits come with patient exposure to ionizing radiation. Low-dose CT (LDCT) reduces this burden, yet dose reduction inevitably lowers the signal-to-noise ratio and introduces noise and structural corruption, especially in anatomically complex regions \cite{waheed2022impact}. LDCT denoising therefore faces a core contradiction: insufficient denoising leaves distracting artifacts, whereas excessive denoising can erase subtle but clinically meaningful structures \cite{kim2019performance}.

Existing LDCT restoration methods approach this trade-off from different angles. LDCT denoising has progressed from model-based filtering and iterative reconstruction to CNN-based, GAN-based, and diffusion-based restoration. CNN-based and GAN-based denoisers are computationally efficient and have substantially improved fidelity and inference speed over earlier approaches, but they may still over-smooth fine detail or introduce unstable texture under severe corruption \cite{redcnn,ldctbenchmark,qae,wganvgg,dugan}. Diffusion models often preserve structure and perceptual quality better by modeling restoration as gradual denoising, yet most current methods operate directly in pixel space and remain expensive on high-resolution CT images, even when sampling is accelerated \cite{ddpm,dndp,yu2024pet,fastddpm}. Latent diffusion is therefore an appealing direction because it can move restoration into a compact domain and improve efficiency, but generic latent diffusion is not directly optimized for CT-specific detail preservation or efficient conditional restoration \cite{ldm,khader2023denoising,petdif}.

Recent work further suggests that restoration quality depends not only on the backbone but also on how conditional information is injected and modulated. Feature-wise conditioning and lightweight gating methods such as FiLM, SE, and UCDIR show that condition integration is an important design choice in restoration and diffusion networks \cite{film,senet,ucdir}. In parallel, frequency-aware restoration methods distinguish stable low-frequency structure from more fragile high-frequency detail, and diffusion studies indicate that higher-frequency guidance may be more effective when introduced progressively across denoising stages \cite{li2018cnn,waveface,fgps,zsfddm,thfnoct}. These observations suggest that efficient latent restoration alone may not be sufficient for LDCT denoising if conditional guidance is not introduced in a detail-aware and timestep-aware manner.

To address these issues, we propose \textbf{TAFG-MAN}, a latent diffusion framework for efficient and high-quality LDCT denoising. The framework uses a perceptually optimized autoencoder to learn a compact latent representation, restores clean latents with a conditional diffusion model, and adopts deterministic DDIM-style sampling \cite{ddim} for efficient inference. To further improve condition utilization during reverse diffusion, we introduce a lightweight \textbf{T}imestep-\textbf{A}daptive \textbf{F}requency-\textbf{G}ated (TAFG) module in the conditioning pathway of the conditional diffusion U-Net. TAFG separates low- and high-frequency condition cues and modulates their contributions according to both the current denoising state and the timestep, while leaving the autoencoder, latent diffusion backbone, and sampler unchanged.

A preliminary version of the latent diffusion backbone, termed \textbf{MAN}, has appeared as an arXiv preprint \cite{fang2025man}. In the present paper, we treat \textbf{MAN} as the base variant of the full framework without TAFG and focus on the complete \textbf{TAFG-MAN} model as the main method. This presentation makes the relationship between the two configurations explicit: MAN provides the shared latent restoration backbone, while TAFG-MAN extends it with timestep-adaptive frequency-gated conditioning. In the experiments, the comparison between MAN and TAFG-MAN therefore serves as a controlled evaluation of the proposed conditioning mechanism.

The main contributions of this work are as follows. First, we propose \textbf{TAFG-MAN}, a latent diffusion framework for LDCT denoising that combines compact perceptual latent modeling, conditional restoration, and efficient deterministic sampling. Second, we design \textbf{TAFG}, a lightweight timestep-adaptive frequency-gated conditioning mechanism that improves how low- and high-frequency condition cues are injected into the latent diffusion U-Net. Third, we provide experiments and ablation studies showing that TAFG-MAN consistently improves over its base variant MAN and offers a favorable balance between restoration quality and inference efficiency compared with representative LDCT denoising baselines.

\section{Method}
\label{sec:method}
\begin{figure}[htb]
\centering
\includegraphics[width=\columnwidth]{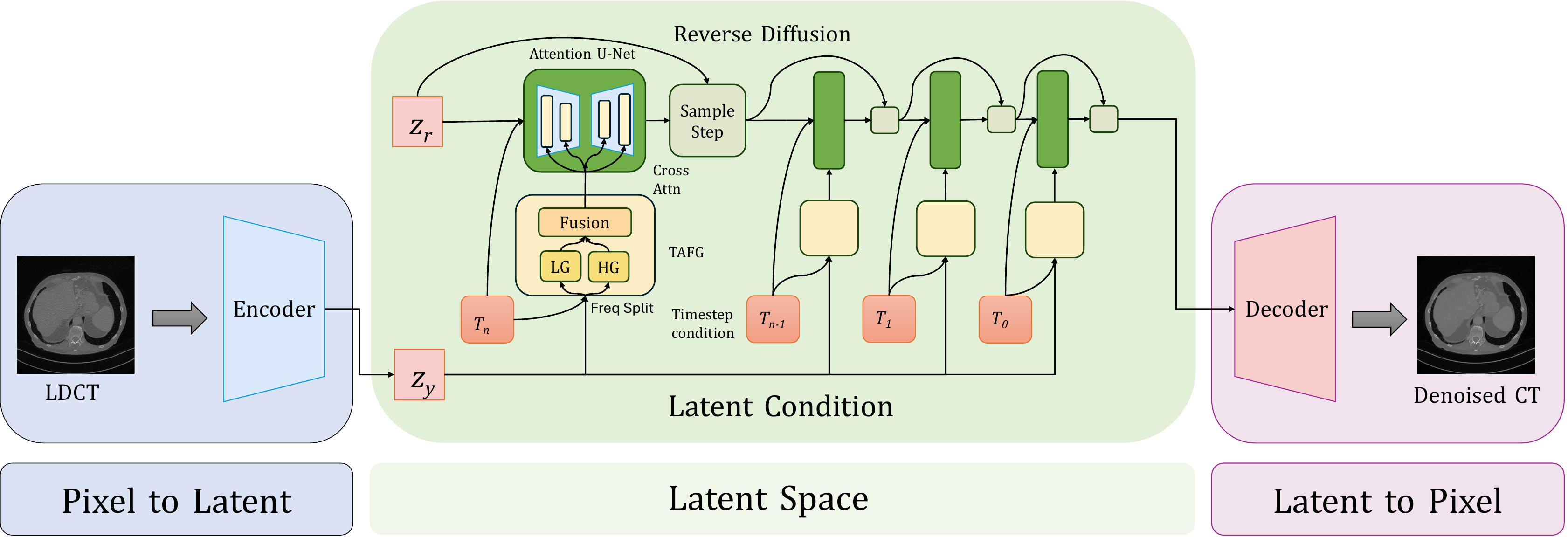}
\caption{Overview of the proposed TAFG-MAN framework for LDCT denoising, LG/HG denote low-/high-frequency gates.}
\label{fig:framework}
\end{figure}

Our goal is to recover a high-quality CT slice $x$ from its low-dose counterpart $y$. Let $\mathcal{E}(\cdot)$ and $\mathcal{D}(\cdot)$ denote the encoder and decoder of the latent autoencoder, respectively. Rather than performing denoising directly in pixel space, MAN formulates restoration in a compact latent space. Given the LDCT input $y$, the encoder $\mathcal{E}(\cdot)$ extracts latent representation $z_y$ as conditioning information, and the conditional latent diffusion model samples a clean latent representation $z_0$ under this condition. The restored CT image is then reconstructed as $x=\mathcal{D}(z_0)$. In practice, the input resolution is $512\times512$, the latent tensor is compressed to $32\times32\times16$, and the observed LDCT image is encoded into both a latent condition and multi-scale condition features that are injected into the latent U-Net through cross-attention. The overall TAFG-MAN framework retains the original two-stage MAN pipeline: (1) a perceptually-optimized autoencoder that maps CT images between pixel and latent domains, and (2) a conditional latent diffusion model that restores clean latent representations using an efficient deterministic sampler. On top of this backbone, TAFG is introduced as a lightweight conditioning upgrade inside the conditional latent diffusion stage. The overall architecture is shown in Figure~\ref{fig:framework}.

\subsection{Perceptually-optimized autoencoder}
\label{ssec:ae}
The first stage of MAN is a customized autoencoder that maps an input image $x \in \mathbb{R}^{H \times W \times 1}$ to a compressed latent representation $z=\mathcal{E}(x) \in \mathbb{R}^{h \times w \times c}$ and reconstructs it back to $\tilde{x}=\mathcal{D}(z)$. The aim is to build a latent space that is simultaneously compact, informative, and well regularized for diffusion.

Architecturally, the encoder follows a U-Net-like design with three downsampling stages that progressively reduce spatial resolution while increasing channel capacity. Each stage contains residual layers with Group Normalization and SiLU activation. To capture longer-range dependencies that are difficult to express with local convolutions alone, self-attention blocks are inserted at the deeper two stages and in the bottleneck. The bottleneck adopts a VAE-like parameterization, producing the latent statistics required for KL regularization. The decoder mirrors the encoder with upsampling and residual refinement blocks, reconstructing the image from latent space while preserving as much local structure as possible.

The autoencoder training objective combines reconstruction accuracy, latent regularity, and perceptual fidelity. We use a weighted combination of pixel-wise MSE and L1 losses to anchor the reconstruction to the original image. Since the latent diffusion stage assumes a reasonably regular latent distribution, we introduce the KL divergence term
\begin{equation}
    \mathcal{L}_{\mathrm{KL}} = \frac{1}{2} \sum_{i=1}^{d} (\mu_i^2 + \sigma_i^2 - \log(\sigma_i^2) - 1),
\end{equation}
which encourages the posterior induced by the encoder to remain close to a standard normal distribution \cite{ldm}. Compared with an aggressively constrained variational formulation, we use KL regularization in a broad and stabilizing manner so that structural detail is not sacrificed for excessive latent smoothness.

To encourage visually faithful reconstructions, we further add a perceptual loss computed from feature activations extracted by a pre-trained VGG-16 network \cite{simonyan2014very}. Denoting the feature map at layer $l$ by $\phi_l(\cdot)$, the perceptual loss is defined as
\begin{equation}
  \mathcal{L}_{\text{PL}} = \sum_{l \in L} \| \phi_l(x) - \phi_l(\tilde{x}) \|_1.
\end{equation}
Although VGG-16 was trained on natural images, previous medical restoration studies have shown that VGG-based feature loss remains useful for preserving structure and texture in medical images \cite{wganvgg}. In our case, this term is particularly important because the latent representation will later serve as the domain on which diffusion operates; if the encoder loses fine structural cues, the subsequent diffusion stage may have limited ability to recover them faithfully.

The overall autoencoder objective is therefore
\begin{equation}
  \mathcal{L}_{\text{AE}} = \mathcal{L}_{\text{pixel}} + \lambda_{\text{KL}}(t)\mathcal{L}_{\text{KL}} + \lambda_{\text{PL}}(t)\mathcal{L}_{\text{PL}},
\end{equation}
where $\mathcal{L}_{\text{pixel}} = \lambda_1 \mathcal{L}_{\text{MSE}} + \lambda_2 \mathcal{L}_{\text{L1}}$. Rather than activating all terms equally from the beginning, we use a linear warm-up schedule during the first half of autoencoder training,
\begin{equation}
\lambda_{\text{KL}}(t)=\lambda_{\text{KL}}^{\max}\min\left(\frac{t}{T_w},1\right), \quad
\lambda_{\text{PL}}(t)=\lambda_{\text{PL}}^{\max}\min\left(\frac{t}{T_w},1\right),
\end{equation}
where $T_w$ denotes the warm-up horizon. Early training therefore focuses on stable reconstruction, after which the regularization and perceptual terms are gradually introduced. This curriculum reduces optimization instability and empirically produces sharper reconstructions with better latent regularity.

\subsection{Conditional latent diffusion}
\label{ssec:ldm}
Once the autoencoder has established a suitable latent space, denoising is performed by a conditional latent diffusion model. The noise prediction network is an attention-enhanced latent U-Net, written as $\epsilon_\theta(z_t,t,c)$, where $z_t$ is the noisy latent at timestep $t$ and $c$ is the conditioning context derived from the LDCT slice $y$. The network consists of residual blocks and deeper spatial transformer blocks equipped with self-attention and cross-attention. Timestep information is encoded through sinusoidal embeddings, and the multi-scale condition features extracted from the LDCT encoder are fused into the denoising trajectory through cross-attention so that restoration remains guided by the observed content.

For the forward process, we adopt the standard diffusion formulation in latent space but use a cosine noise schedule rather than the original linear schedule. Specifically,
\begin{equation}
    \bar{\alpha}_t = \frac{\cos^2\left( \frac{(t / T + \epsilon)}{1 + \epsilon} \cdot \frac{\pi}{2} \right)}{\cos^2\left( \frac{\epsilon}{1 + \epsilon} \cdot \frac{\pi}{2} \right)}, \quad \text{with } \epsilon = 0.008,
\end{equation}
\begin{equation}
    \beta_t = 1 - \frac{\bar{\alpha}_{t}}{\bar{\alpha}_{t-1}}.
\end{equation}
We set the total diffusion horizon to $T=500$ during training. This schedule allocates noise more smoothly across timesteps and is commonly found to improve denoising stability in diffusion training \cite{nichol2021improved}.

The reverse process is where MAN departs from conventional latent diffusion most clearly. Instead of the stochastic DDPM-style sampler, we adopt a deterministic DDIM-style sampler \cite{ddim}. For restoration tasks, this choice is desirable because the target output for a given input is not arbitrary: different runs should ideally converge to the same clean solution. The DDIM-style update used in MAN is
\begin{equation}
\begin{aligned}
    z_{t-1} &= \sqrt{\bar{\alpha}_{t-1}}\left(\frac{z_t - \sqrt{1-\bar{\alpha}_t}\epsilon_\theta(z_t, t, c)}{\sqrt{\bar{\alpha}_t}}\right) \\
    &\quad + \sqrt{1-\bar{\alpha}_{t-1}}\epsilon_\theta(z_t, t, c).
\end{aligned}
\end{equation}
At inference time, we use a reduced DDIM step budget (30 steps in our main setting), which is a small fraction of the full training horizon. Compared with stochastic reverse diffusion, this deterministic path substantially shortens inference and yields more stable outputs.

The latent U-Net is trained with the standard noise-prediction objective
\begin{equation}
  \mathcal{L}_{\text{LDM}} = \mathbb{E}_{z_0, \epsilon, t} \left[ \| \epsilon - \epsilon_\theta(\sqrt{\bar{\alpha}_t}z_0 + \sqrt{1-\bar{\alpha}_t}\epsilon, t, c) \|^2 \right],
\end{equation}
where $t$ is sampled uniformly from $\{1,\ldots,T\}$. Here, the model learns to predict the synthetic Gaussian noise added during forward diffusion. After reverse denoising produces $\tilde{z}_0$, the decoder reconstructs the final image $\tilde{x}=\mathcal{D}(\tilde{z}_0)$.

\subsection{Timestep-Adaptive Frequency-Gated Conditioning}
\label{ssec:tafg}
While the MAN backbone already uses multi-scale conditioning features to guide latent diffusion, the original condition injection remains relatively uniform across diffusion steps. Recent restoration studies suggest that the way condition information is integrated can significantly affect the final reconstruction quality rather than serving as a minor architectural detail \cite{film,senet,ucdir}. In LDCT denoising, however, the condition feature at a given scale may simultaneously contain stable low-frequency structure, useful high-frequency edges, and noise-contaminated high-frequency responses. Injecting these components with the same strength can make it harder to balance aggressive denoising against fine-detail preservation. To address this issue, we introduce a lightweight \textbf{T}imestep-\textbf{A}daptive \textbf{F}requency-\textbf{G}ated (TAFG) conditioning module that operates only on the condition path before cross-attention.

Consider a condition feature map $c_l \in \mathbb{R}^{H_l \times W_l \times C_l}$ at a selected latent U-Net scale $l$. We first decompose it into low- and high-frequency components:
\begin{equation}
c_l^{\mathrm{low}} = P(c_l), \qquad c_l^{\mathrm{high}} = c_l - c_l^{\mathrm{low}},
\end{equation}
where $P(\cdot)$ is a lightweight learnable low-pass operator. This decomposition is motivated by prior frequency-aware restoration studies showing that low- and high-frequency components often contribute differently to structure preservation and detail recovery \cite{li2018cnn,waveface}. In our implementation, $P(\cdot)$ is realized by a depthwise low-pass convolution, which provides a simple local frequency split without introducing a heavy FFT- or wavelet-based branch.

To modulate these two components according to the current denoising state, TAFG predicts separate low- and high-frequency gates from the current U-Net feature $f_l$ and timestep embedding $e_t$. This design follows the general intuition of lightweight feature-wise modulation and channel gating, but specializes it to frequency-aware conditional guidance in latent diffusion \cite{film,senet}. Specifically, we first apply global average pooling to $f_l$, concatenate the pooled feature with $e_t$, and feed the result into a lightweight MLP:
\begin{equation}
\left[g_l^{\mathrm{low}}, g_l^{\mathrm{high}}\right] =
\sigma\!\left(\mathrm{MLP}\left([\mathrm{GAP}(f_l), e_t]\right)\right),
\end{equation}
where $\sigma(\cdot)$ denotes the sigmoid function. The two gates control how strongly the low- and high-frequency condition components should be injected at the current denoising step.

We further introduce a timestep-dependent release factor for the high-frequency branch. Let $\hat{t}=t/(T-1)$ denote the normalized timestep, where $\hat{t}=1$ corresponds to the noisiest state and $\hat{t}=0$ to the cleanest state. The release factor is defined as
\begin{equation}
r(t) = (1-\hat{t})^{\gamma},
\end{equation}
where $\gamma$ is a nonnegative exponent controlling how gradually high-frequency guidance is released. This design is inspired by recent diffusion restoration studies suggesting that higher-frequency information can be introduced more effectively in a progressive, timestep-dependent manner \cite{fgps,zsfddm,thfnoct}. It suppresses overly aggressive high-frequency injection at early denoising stages and progressively allows more detail-oriented guidance as reverse diffusion approaches the clean image manifold.

The final modulated condition feature is therefore
\begin{equation}
\tilde{c}_l = g_l^{\mathrm{low}} \odot c_l^{\mathrm{low}} + r(t)\, g_l^{\mathrm{high}} \odot c_l^{\mathrm{high}},
\end{equation}
where $\odot$ denotes element-wise multiplication. The modulated feature $\tilde{c}_l$ is then reshaped into context tokens and fed into the original cross-attention layers. In this way, TAFG improves the conditioning pathway without changing the pretrained autoencoder, the latent diffusion backbone, or the DDIM-style sampler. In our implementation, the module is inserted at selected scales of the latent U-Net, allowing the model to combine stable structural guidance at earlier stages with progressively released high-frequency detail in later stages of denoising.

\section{Experiments}
\label{sec:exp}
\subsection{Experimental setup}
\label{ssec:setup}
We evaluate MAN on a subset of the LDCT and Projection dataset \cite{mccollough2020low}. Following prior LDCT studies \cite{redcnn,ldctbenchmark,dndp,fastddpm}, we use 1~mm abdominal and chest CT slices of size $512\times512$ with D45 kernel. Data are
split in 70\%/10\%/20\% training/validation/test set in a patient-wise split following the split protocol of LDCT-Bench\cite{ldctbenchmark}. Performance is assessed using PSNR, SSIM, and LPIPS \cite{zhang2018unreasonable}; for LPIPS, each single-channel CT slice is replicated to three channels after identical clipping and normalization to $[-1,1]$. We compare against BM3D \cite{bm3d}, RED-CNN \cite{redcnn}, Q-AE \cite{qae}, WGAN-VGG \cite{wganvgg}, DU-GAN \cite{dugan}, DDPM \cite{ddpm}, Fast-DDPM \cite{fastddpm}, and Dn-Dp \cite{dndp}, covering conventional, CNN-based, GAN-based, and diffusion-based LDCT denoisers.

Our method is implemented in PyTorch 2.8 with CUDA 12.8 and trained on two NVIDIA RTX 4080 Super GPUs. The autoencoder is pre-trained for 180000 iterations with a simulated batch size of 16, after which its weights are frozen and the latent diffusion model is trained for 240000 iterations on the learned latent space using mixed precision. The latent representation has size $32\times32\times16$; diffusion uses a cosine schedule with $T=500$ during training \cite{nichol2021improved}, and inference uses a deterministic 30-step DDIM-style \cite{ddim} quick sampler. We use Adam with learning rate $10^{-4}$ and pixel reconstruction weights $\lambda_1=1$ and $\lambda_2=0.5$, while the perceptual and KL terms are warmed up gradually during early autoencoder training \cite{ldm,fastddpm}. 

To keep comparisons as fair as possible, all methods are evaluated under the same HU clipping range, patient-wise test split, and metric pipeline whenever unified reproduction is feasible, and runtime is measured as end-to-end denoising time on the same hardware platform. Since BM3D \cite{bm3d} is a CPU-based method, its inference time is reported on the same CPU, while all learning-based methods are evaluated on the same GPU.

\subsection{Quantitative comparison}
\label{ssec:quant}
The main quantitative results are summarized in Table~\ref{tab:comparison}. The evaluation is performed at each method's native output resolution to avoid interpolation-induced bias. Fast-DDPM natively produces 256×256 outputs, so its metrics are computed against ground truth at the same resolution, whereas all other methods are evaluated at their native 512×512 resolution. Overall, both MAN and TAFG-MAN are highly competitive across distortion, structure, perceptual quality, and efficiency, showing the benefit of combining a CT-oriented perceptual autoencoder with latent diffusion and deterministic DDIM-style sampling. Compared with the broader baselines, RED-CNN remains competitive in PSNR but is clearly weaker in LPIPS, and pixel-space diffusion methods such as DDPM and Dn-Dp are either slower or less balanced overall. Fast-DDPM provides high denoising performance, but slightly slower in inference speed. These results indicate that MAN establishes the main quality-efficiency advantage of the framework, while TAFG further improves structural fidelity and perceptual realism through more effective conditional modulation.

\begin{table}[htb]
    \centering
    \small
    \begin{tabular}{l|cccc}
        \hline
        \textbf{Method} & \textbf{PSNR} $\uparrow$ & \textbf{SSIM} $\uparrow$ & \textbf{LPIPS} $\downarrow$ & \textbf{Time} $\downarrow$\\
        \hline
        BM3D\cite{bm3d} & 30.09 & 0.660 & 0.370 & 21.0m \\
        RED-CNN\cite{redcnn} & 31.38 & 0.693 & 0.272 & 3.7s \\
        Q-AE\cite{qae} & 30.19 & 0.652 & 0.246 & 0.6s \\
        WGAN-VGG\cite{wganvgg} & 27.03 & 0.533 & 0.300 & 0.3s \\
        DU-GAN\cite{dugan} & 30.35 & 0.683 & 0.211 & 3.8s \\
        DDPM\cite{ddpm} & 29.44 & 0.655 & 0.312 & 47m \\
        Fast-DDPM\cite{fastddpm} & 31.18 & 0.701 & 0.251 & 31.5s \\
        Dn-Dp\cite{dndp} & 29.02 & 0.623 & 0.211 & 18.7m \\
        \textbf{MAN (Ours)} & 31.16 & 0.703 & 0.119 & 18.8s \\
        \textbf{TAFG-MAN (Proposed)} & \textbf{31.43} & \textbf{0.722} & \textbf{0.100} & 18.7s \\
        \hline
    \end{tabular}
    \caption{Quantitative comparison of CT denoising methods. Higher PSNR and SSIM indicate better distortion/structure preservation, while lower LPIPS indicates better perceptual fidelity.}
    \label{tab:comparison}
\end{table}

\subsection{Qualitative results}
\label{ssec:qual}
\begin{figure}[htb]
\centering
\includegraphics[width=\columnwidth]{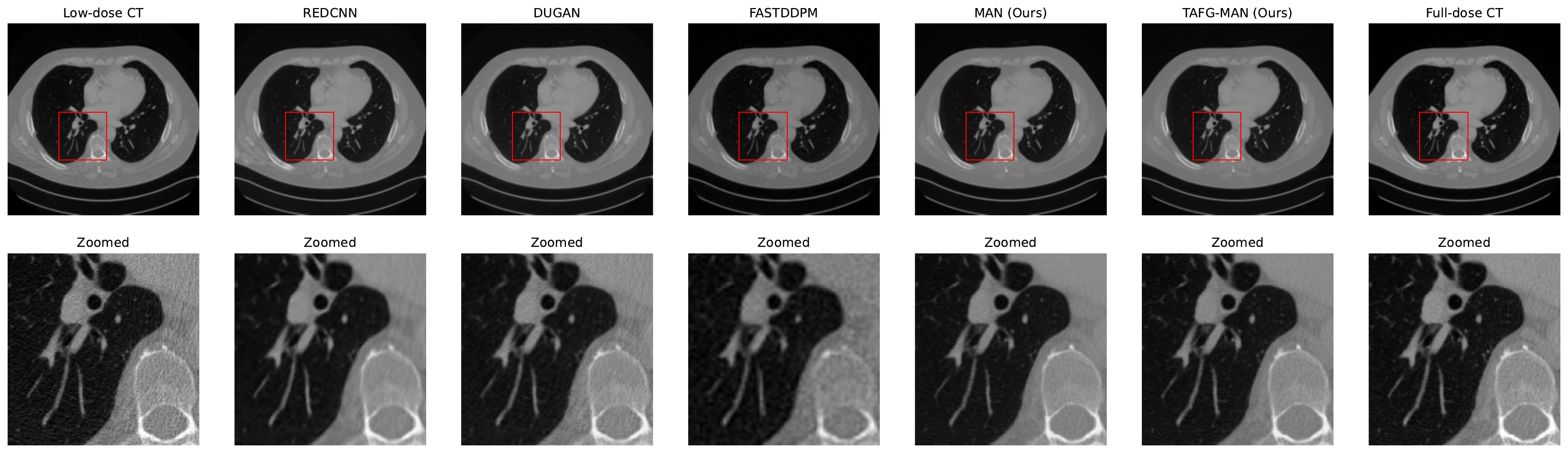}
\vspace{0.5em}
\includegraphics[width=\columnwidth]{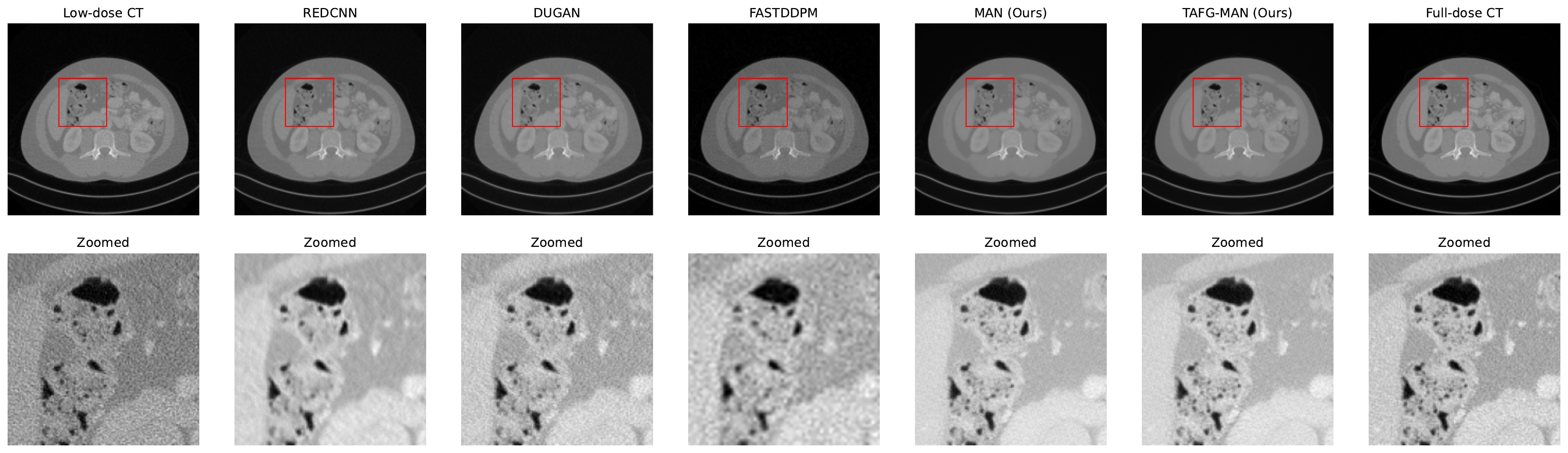}
\caption{Visual comparison of denoising results on two representative test slices. For visualization only, the output of Fast-DDPM, originally $256\times256$ in its native setting, was upscaled to $512\times512$ using bicubic interpolation.}
\label{fig:visual}
\end{figure}

Figure~\ref{fig:visual} provides representative visual comparisons that are consistent with the quantitative results. RED-CNN suppresses noise effectively but tends to smooth away subtle structural variation, whereas DU-GAN preserves texture more aggressively at the cost of residual noise or local artifacts. Fast-DDPM demonstrates generally good visual performance at its native resolution, but does not align with mainstream resolution, implying additional post-processing costs. MAN already provides a stronger balance between suppression and detail retention through its perceptually optimized latent representation and stable deterministic restoration process, and TAFG-MAN further improves this balance by recovering cleaner local structure, sharper boundaries, and finer vessels with fewer over-smoothed regions. The visual results therefore support the same conclusion as Table~\ref{tab:comparison}: the main robustness comes from the MAN backbone, while TAFG improves the use of conditional guidance for detail preservation.

\subsection{Ablation study}
\label{ssec:ablation}
To assess the contribution of the major design components, we conduct the ablation study summarized in Table~\ref{tab:ablation}. Starting from a minimal latent diffusion baseline with a simple autoencoder and a conventional DDPM-style sampler, we progressively introduce the perceptually-optimized autoencoder (P-AE), the DDIM-style quick sampler (Q-SP), and TAFG. The perceptual autoencoder alone mainly improves SSIM and LPIPS, indicating that better latent representations already benefit restoration quality. Replacing the DDPM-style sampler with Q-SP yields the largest efficiency gain, reducing inference from minutes to seconds while also improving all three image-quality metrics. Adding TAFG on top of the MAN configuration further improves PSNR, SSIM, and LPIPS with almost no runtime cost, and the full combination achieves the best overall result. Taken together, the ablation shows that latent representation quality, efficient deterministic sampling, and timestep-adaptive frequency-aware conditioning contribute complementary gains rather than redundant ones.

\begin{table}[htb]
    \centering
    \small
    \begin{tabular}{ccc|cccc}
        \hline
        \multicolumn{3}{c|}{\textbf{Components}} & \multicolumn{4}{c}{\textbf{Performance}} \\
        \hline
        \textbf{P-AE} & \textbf{Q-SP} & \textbf{TAFG} & \textbf{PSNR} $\uparrow$ & \textbf{SSIM} $\uparrow$ & \textbf{LPIPS} $\downarrow$ & \textbf{Time} $\downarrow$\\
        \hline
        \texttimes & \texttimes & \texttimes & 29.72 & 0.676 & 0.156 & 3.29m\\
        \checkmark & \texttimes & \texttimes & 29.68 & 0.682 & 0.153 & 3.35m\\
        \texttimes & \checkmark & \texttimes & 30.93 & 0.695 & 0.132 & 18.98s\\
        \checkmark & \checkmark & \texttimes & 31.13 & 0.708 & 0.118 & 18.87s\\
        \checkmark & \checkmark & \checkmark & \textbf{31.43} & \textbf{0.722} & \textbf{0.100} & \textbf{18.71s}\\
        \hline
    \end{tabular}
    \caption{Ablation study of the proposed perceptual autoencoder (P-AE), quick sampler (Q-SP), and TAFG conditioning module.}
    \label{tab:ablation}
\end{table}

\section{Conclusion and Limitations}
\label{sec:concl}
We presented TAFG-MAN, a latent diffusion framework for efficient and high-quality LDCT denoising. The framework combines a CT-oriented perceptual autoencoder, conditional latent diffusion, deterministic DDIM-style sampling, and timestep-adaptive frequency-gated condition modeling. Extensive experiments and ablation studies show that TAFG-MAN improves over its base variant and remains competitive with or superior to representative CNN, GAN, and pixel-space diffusion baselines, indicating that efficient latent diffusion with adaptive condition modeling is a practical direction for medical image restoration.

This work still has several limitations. First, the experiments are conducted on a specific LDCT subset, so broader validation on additional anatomies, scanners, and acquisition protocols is still needed. Second, the latent diffusion process relies on a simplified Gaussian-style corruption model, whereas real CT noise can be more complex and acquisition dependent. Third, although TAFG improves the balance between noise suppression and detail preservation, it remains a lightweight local conditioning mechanism and does not explicitly model longer-range anatomical priors. Finally, the current study does not include a radiologist reader evaluation, so the clinical significance of the perceptual improvements remains to be further assessed.
%
%
%
\bibliographystyle{splncs04}
\bibliography{cite}
%




\end{document}